\documentclass{article} 
\usepackage{iclr2026_conference,times}

\usepackage{amsmath,amsfonts,bm}

\def\eqref#1{equation~\ref{#1}}

\def\1{\bm{1}}

\DeclareMathAlphabet{\mathsfit}{\encodingdefault}{\sfdefault}{m}{sl}
\SetMathAlphabet{\mathsfit}{bold}{\encodingdefault}{\sfdefault}{bx}{n}

\usepackage{url}
\usepackage{hyperref}
\usepackage{bbding}
\usepackage{amsfonts}       
\usepackage{nicefrac}       
\usepackage{microtype}      
\usepackage{xcolor}         
\usepackage{multirow}
\usepackage{colortbl}
\usepackage{algorithm}
\usepackage{cleveref}
\usepackage{algpseudocode}
\usepackage{graphicx}
\usepackage{booktabs}
\usepackage{amsmath}
\usepackage{wrapfig}
\usepackage{subcaption}
\usepackage{utfsym}
\usepackage{etoc}
\crefname{section}{Sec.}{Secs.}
\Crefname{section}{Section}{Sections}
\crefname{appendix}{Appendix}{Appendices}
\Crefname{appendix}{Appendix}{Appendices}
\crefname{table}{Tab.}{Tabs.}
\crefname{figure}{Fig.}{Figs.}
\crefname{equation}{Eq.}{Eqs.}
\crefname{theorem}{Thm.}{Thms.}
\crefname{lemma}{Lem.}{Lems.}
\crefname{remark}{Rem.}{Rems.}
\crefname{corollary}{Cor.}{Cors.}
\crefname{algorithm}{Alg.}{Algs.}

\definecolor{mypup}{RGB}{66,133,244}
\definecolor{mypup}{RGB}{51,168,83}
\definecolor{myyellow}{RGB}{251,188,3}
\definecolor{myred}{RGB}{234,67,53}
\definecolor{mygrey}{RGB}{95,99,104}
\definecolor{mypup}{RGB}{153,0,204}

\title{Any-to-Bokeh: Arbitrary-Subject Video Refocusing with Video Diffusion Model}

\author{
\begin{minipage}{\textwidth}
\centering
Yang Yang$^{1,2*}$ \quad
Siming Zheng$^{2*}$ \quad
Qirui Yang$^{2}$ \quad
Jinwei Chen$^{2}$ \quad
Boxi Wu$^{{1\dag}}$
\end{minipage}\\[4pt]  
\begin{minipage}{\textwidth}
\centering
\textbf{Xiaofei He}$^1$ \quad
\textbf{Deng Cai}$^1$ \quad
\textbf{Bo Li}$^2$ \quad
\textbf{Peng-Tao Jiang}$^{2\dag}$
\end{minipage}\\[3pt]
\begin{minipage}{\textwidth}
\centering
$^1$Zhejiang University \quad
$^2$vivo Mobile Communication Co., Ltd \\
\end{minipage}\\[3pt]
\texttt{Project Page: \url{https://vivocameraresearch.github.io/any2bokeh/}}
}

\iclrfinalcopy 
\begin{document}

\maketitle

\begin{figure}[ht]
\vspace{-16pt}
\begin{center}
	\includegraphics[width=1\linewidth]{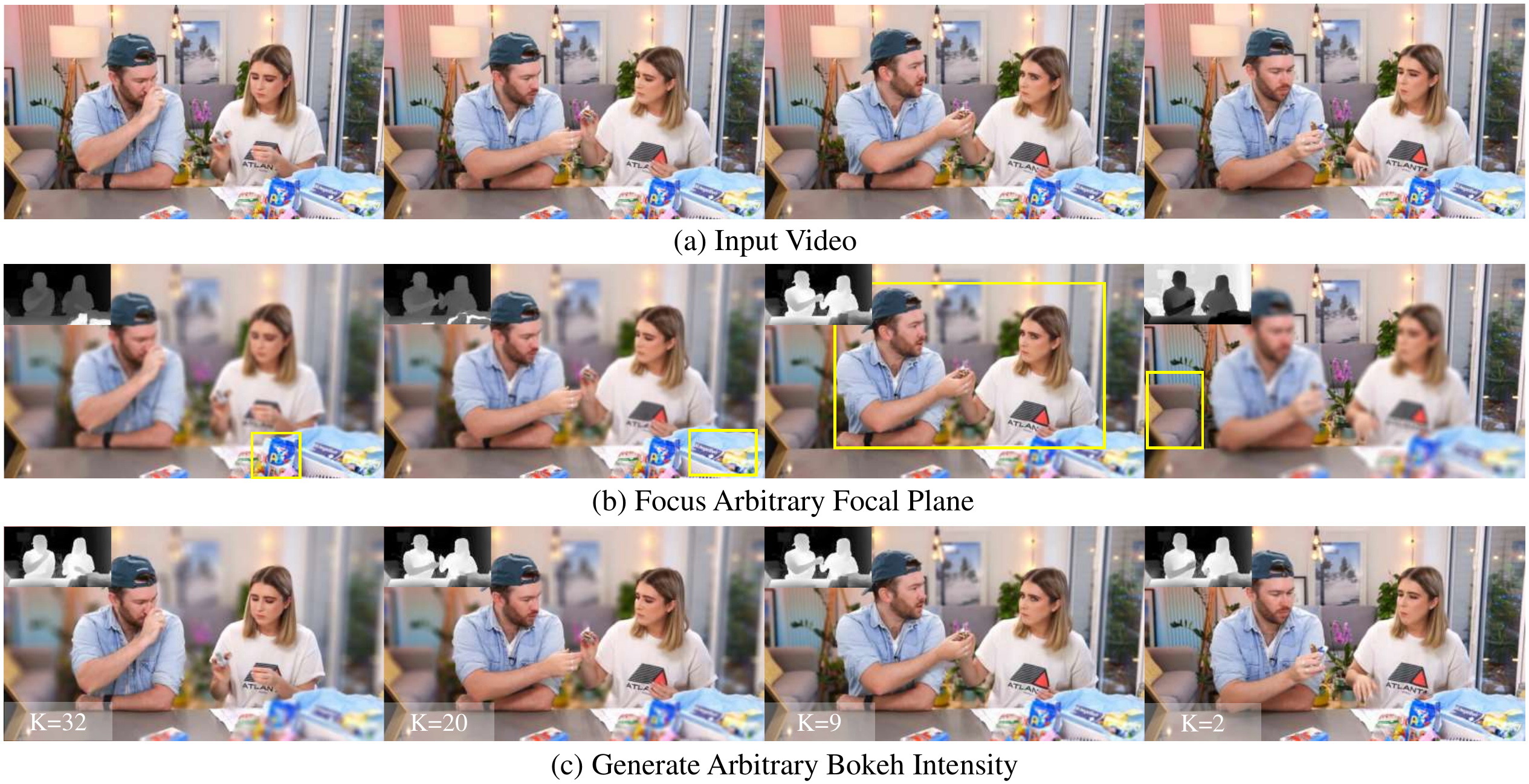}
\end{center}
\vspace{-5mm}
\caption{
\textbf{Any-to-Bokeh enables users to customize the focal plane and adjust bokeh intensity}. The yellow box indicates the focal plane, and the grayscale values in the image represent the distance to the focal plane, with higher values indicating closer proximity. $K$ represents the bokeh intensity.
}
\label{fig:teaser}
\end{figure}

\begin{abstract}

Diffusion models have recently emerged as powerful tools for camera simulation, enabling both geometric transformations and realistic optical effects. Among these, image-based bokeh rendering has shown promising results, but diffusion for video bokeh remains unexplored.
Existing image-based methods are plagued by temporal flickering and inconsistent blur transitions, while current video editing methods lack explicit control over the focus plane and bokeh intensity.
These issues limit their applicability for controllable video bokeh.
In this work, we propose a one-step diffusion framework for generating temporally coherent, depth-aware video bokeh rendering. 
The framework employs a multi-plane image (MPI) representation adapted to the focal plane to condition the video diffusion model, thereby enabling it to exploit strong 3D priors from pretrained backbones.
To further enhance temporal stability, depth robustness, and detail preservation, we introduce a progressive training strategy.
Experiments on synthetic and real-world benchmarks demonstrate superior temporal coherence, spatial accuracy, and controllability, outperforming prior baselines. 
This work represents the first dedicated diffusion framework for video bokeh generation, establishing a new baseline for temporally coherent and controllable depth-of-field effects. Code will be made publicly available.

\end{abstract}

\vspace{-2mm}
\section{Introduction}
\label{sec: introduction}
\vspace{-2mm}

Recent advances in diffusion models have significantly advanced camera simulation, enabling controllable geometric transformations such as lens movement, zooming, and panning~\citep{teng2023drag, shi2024dragdiffusion, wu2024draganything, yin2023dragnuwa}. 
These developments demonstrate their strong potential for reproducing realistic optical effects beyond traditional rendering pipelines.

Building on these advances, diffusion-based methods have recently been extended to image bokeh rendering~\citep{fortes2025bokeh, zhu2025bokehdiff}. 
By leveraging generative priors, these methods achieve visually convincing blur transitions and highlight the feasibility of applying diffusion models to depth-aware, image-level optical effects. 
However, these methods focus on image inputs, with limited temporal modeling, so their applicability to videos is constrained.

Compared to image-based bokeh, video bokeh poses additional challenges remains limited in terms of temporal coherence and controllability.
The limitations are two-fold.
(i) Frame-by-frame extensions of image methods~\citep{sheng2024dr,peng2022mpib,peng2022bokehme,deeplens2018,zhu2025bokehdiff} lack explicit temporal modeling and rely on imperfect depth estimates. Together with stochastic noise accumulation in multi-step diffusion, this often causes \textbf{temporal flickering} and unstable blur boundaries, degrading video fidelity.
(ii) Although recent video generative models~\citep{blattmann2023stable} provide strong 3D priors and cross-frame consistency, their bokeh-like effects are largely implicit and lack explicit control over the \textbf{focal plane} and \textbf{bokeh intensity}, limiting realistic and controllable video bokeh. 
These limitations highlight the need for a dedicated framework for temporally coherent, depth-aware video bokeh.
To overcome these limitations, we introduce a new framework that explicitly models scene geometry and leverages strong pre-trained priors to achieve temporally coherent, controllable video bokeh.
A central challenge in video bokeh is ensuring smooth blur transitions across depth, particularly near object boundaries, where inaccurate depth often causes artifacts. To address this, we adopt a multi-plane image (MPI) representation, which provides a compact yet explicit encoding of scene geometry. 
By constructing MPI layers with a progressively widening disparity sampling function, our method captures fine details for in-focus regions while allocating coarser representation to out-of-focus areas, thereby enabling accurate and stable blur transitions across depth boundaries.

Building on this representation, we condition a one-step video diffusion model on MPI layers, guiding the network to synthesize depth-aware bokeh effects that align with subject contours even in cluttered scenes.
Unlike prior methods trained from scratch~\citep{luo2024video, zhang2019synthetic}, our framework leverages large-scale pre-trained video diffusion models~\citep{blattmann2023stable}, whose strong 3D priors provide robust generalization across diverse scenarios and improve structural consistency.
To further enhance temporal stability and detail preservation, we design a three-stage progressive training strategy that (i) establishes accurate geometric guidance via MPI-based spatial and temporal modules, (ii) improves robustness and reduces flickering through extended temporal windows with data perturbations, and (iii) refines subject details using a VAE-based enhancement module.
Together, these components form a unified framework that directly addresses the core challenges of temporal coherence, geometric accuracy, and visual quality in video bokeh generation.

Extensive experiments on both synthetic and real-world benchmarks demonstrate that our framework produces high-quality, controllable video bokeh and significantly outperforms prior baselines. By supporting rendering from arbitrary video inputs, our method makes video bokeh generation both practical and generalizable, opening new possibilities for content creation, cinematic editing, and mobile post-processing. Our main contributions are as follows:
\vspace{-1mm}
\begin{itemize}
\item We propose the first one-step diffusion framework for controllable video bokeh, introducing an MPI-guided conditioning mechanism that injects explicit scene geometry to enable spatially accurate, depth-aware blur synthesis. 
\item We design a progressive training strategy that improves temporal stability and detail fidelity, yielding realistic and controllable video bokeh across diverse scenarios.
\item Our approach achieves state-of-the-art in-focus boundary fidelity and temporal consistency, producing controllable video bokeh for arbitrary input videos with explicit control over the focus plane and bokeh intensity.
\end{itemize}

\section{Related work}
\vspace{-0.15cm}
\subsection{Camera Simulation Diffusion Models}
\vspace{-0.1cm}
\textbf{Reference Guidance Models}. 
Recent advancements in camera simulation have demonstrated the potential of leveraging reference videos to guide the motion and behavior of virtual cameras.
A line of work~\citep{zhao2024motiondirector,guo2023animatediff} encodes motion cues from reference videos via LoRA~\citep{hu2022lora}, enabling the diffusion model to replicate specific camera behaviors observed in the training set. In contrast, MotionClone~\citep{ling2025motionclone} proposes a training-free approach, directly extracting motion patterns from reference videos via spatial and temporal attention modules. Other approaches~\citep{teng2023drag, shi2024dragdiffusion, wu2024draganything, yin2023dragnuwa} require users to draw reference points to guide lens adjustments, offering more direct control over camera transformations.

\textbf{Optical Effects with Diffusion.}
While earlier diffusion-based camera simulation methods mainly focused on motion modeling, recent studies have expanded to simulating optical effects such as bokeh. 
For example, BokehDiffusion~\citep{fortes2025bokeh} and Generative Photography~\citep{Yuan_2025_GenPhoto} generate high-quality bokeh effects from 2D images, capturing realistic defocus effects. These methods demonstrate the potential of diffusion models for simulating non-geometric camera behavior, but they are restricted to static images and lack temporal modeling, which limits their applicability to video tasks. In contrast, our approach integrates optical effect simulation with pre-trained priors, enabling temporally consistent and controllable bokeh effects for arbitrary video inputs. This work, therefore, bridges the gap between image-based bokeh rendering and the video domain.

\subsection{Computational Bokeh}
\textbf{Training-Free Methods}. 
Early computational bokeh methods are training-free and rely on physically-based or image-based heuristics. Ray tracing~\citep{pharr2023physically,potmesil1981lens,akenine2019real} produces realistic defocus effects, but requires full 3D geometry and is computationally expensive, which limits its practical applicability. Depth-based methods~\citep{wadhwa2018synthetic,yang2016virtual} apply scattering or gathering kernels to create spatially varying blur using estimated depth. Matting-based methods~\citep{shen2016automatic,shen2016deep} blur the background based on foreground masks. Although these methods do not require training, they often suffer from artifacts due to inaccurate depth estimation or segmentation errors, leading to unnatural blur transitions.

\textbf{Learning-based Methods}.
More recent approaches adopt machine learning to improve bokeh synthesis. For example, BokehMe~\citep{peng2022bokehme} refines depth-based blur using classical methods~\citep{wadhwa2018synthetic} along with neural networks, while models based on Multiplane Images (MPI)\citep{zhou2018stereo}\citep{peng2022mpib, busam2019sterefo, sheng2024dr} decompose scenes into layered representations to render depth-aware effects. Other works explore adaptive kernels or light field approximations~\citep{Srinivasan_2018_CVPR, kaneko2021unsupervised} to enhance bokeh effects.
DeepLens~\citep{deeplens2018} introduces a depth estimation network using depth estimation and foreground segmentation data to enhance the perception of foreground edges
End-to-end models trained on paired all-in-focus and bokeh datasets~\citep{ignatov2020rendering,seizinger2025bokehlicious,dutta2021stacked} show promise but are limited by dataset biases and fixed parameter simulation. They generally lack the flexibility to control focal planes or simulate custom bokeh intensities.

While image-based bokeh synthesis is well-studied, extending it to videos is challenging. Naive frame-wise methods often introduce temporal flickering and inconsistency. Our work addresses this by a novel MPI-guided conditioning mechanism and leveraging pre-trained video diffusion models for consistent spatial and temporal bokeh rendering.

\section{Method}

\begin{figure}[tb]
  \centering
  \includegraphics[width=\linewidth]{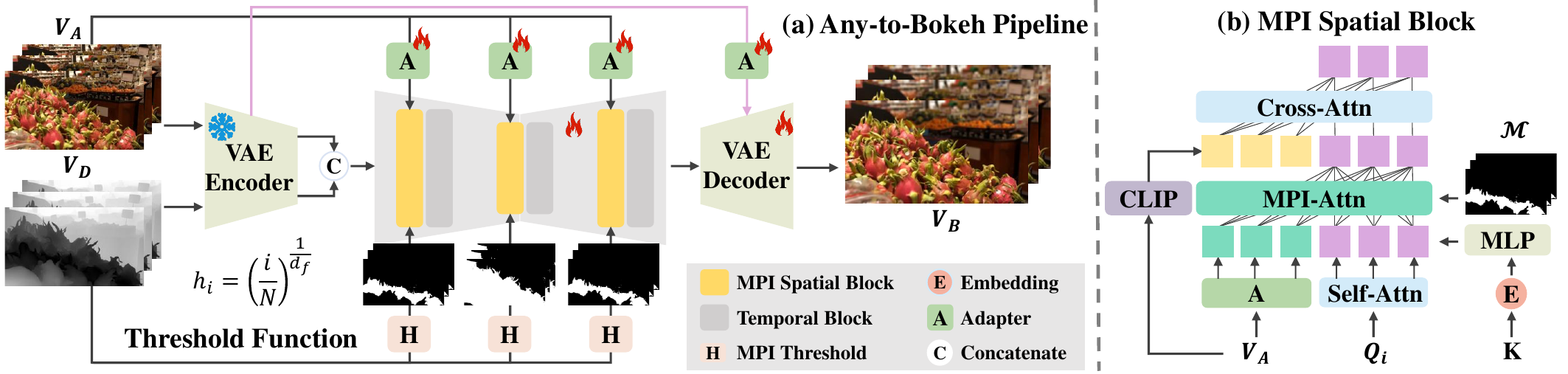}
    \vspace{-6mm}
  \caption{
   Two key components of Any-to-Bokeh. (a) One-step video bokeh pipeline: receives input of any video and disparity relative to the focal plane to perform the bokeh effect. (b) MPI spatial block: uses the MPI mask $\mathcal{M}$ to prompt MPI spatial block to guide bokeh rendering. The user-defined blur strength $K$ is injected through embedding.
  }
  \label{fig: pipeline}
  \vspace{-6mm}
\end{figure}

An overview of our pipeline is shown in \cref{fig: pipeline}(a). Our framework performs video bokeh rendering in a one-step diffusion scheme, guided by MPI priors. The design comprises three components: (i) a focal-plane–adapted MPI representation that provides geometry-aware spatial priors, (ii) a one-step video diffusion backbone conditioned on MPI-derived and user prompts that generates controllable and meticulous video bokeh rendering, and (iii) a progressive training strategy that improves temporal stability and detail fidelity.

\subsection{Focal-Plane–Adapted MPI Representation}
A central challenge in video bokeh is achieving realistic blur transitions, especially near object boundaries, where depth discontinuities often cause visible artifacts.
Prior works~\citep{peng2022mpib,busam2019sterefo} discretize the scene into front-to-back layers with fixed depth values. While effective for simple rendering, such fixed discretizations are poorly aligned with the optics of bokeh: blur radius varies nonlinearly with respect to the focal plane, and equal-depth intervals fail to capture these variations. This mismatch often produces inaccurate blur transitions and artifacts around fine structures.
Therefore, we introduce a focal-plane-adapted MPI representation that explicitly drives the geometry prior relevant to bokeh formation. This adaptive scheme yields more accurate depth-dependent blur and ensures smooth transitions across subject contours. Specifically, the optics of the circle of confusion (CoC) relates the blur radius $r$ to disparity:
\begin{equation}
\label{equ:coc}
r = K\left|\frac{1}{z}-\frac{1}{z_f}\right| = K|d-d_f|,
\end{equation}
where $K$ is the user-controlled blur strength; $z$ and $d$ denote depth and disparity, respectively; and $z_f$ and $d_f$ are the corresponding focal-plane values. Since $r$ changes more rapidly near shallow focal planes and more slowly in distant regions, we sample disparity more finely near $d_f$ and more coarsely farther away.
Concretely, for each video frame, we predict a disparity map $d$ using a pre-trained depth estimator and define $N$ thresholds with a threshold function:
\begin{equation}
\label{equ:threshold}
h_i=\Big(\tfrac{i}{N}\Big)^{\frac{1}{d_f}},\quad i=1,2,\dots,N-1,
\end{equation}
where $1/d_f\in(0,1]$ acts as a factor, yielding finer sampling for shallow focus. Pixels are assigned to MPI layers using these thresholds, producing a focal-plane–adapted MPI mask
\begin{equation}
\mathcal{M} = \{ m_i \mid |d(m_i) - d_f| < h_i \},
\end{equation}
which highlights regions close to the focus and provides finer granularity around depth discontinuities. In addition, we compute a normalized disparity-difference map $V_D$ from $d$ and $d_f$ (an implicit CoC proxy per \cref{equ:coc}). Together, $\mathcal{M}$, $V_D$, and $K$ constitute the geometry prior and user-guided controls that will condition the generator. Unlike fixed front-to-back discretizations~\citep{peng2022mpib,busam2019sterefo}, our MPI is defined \emph{relative to the focal plane}, enabling focus-aware sampling and accurate boundary transitions. 

\subsection{One-Step Video Bokeh Diffusion on MPI Priors}
To avoid the temporal instability across frames caused by repeated iterations of the multi-step diffusion model and the high computational cost of multi-step inference, we formulate the generation of video bokeh as a one-step diffusion guided by the MPI prior. Building on Stable Video Diffusion (SVD)~\citep{blattmann2023stable}, we remove stochastic sampling and adopt a one-step U-Net that directly predicts the output frames.

The backbone is conditioned on three explicit signals: i) \textbf{normalized disparity difference} $V_D$ (focus proximity); ii) \textbf{blur strength} $K$ (bokeh intensity); iii) \textbf{focal-plane–adapted MPI mask} $\mathcal{M}$ (geometry prior).
This conditioning provides precise control over focus placement and bokeh intensity, while promoting frame-to-frame consistency.

\begin{figure}[tb]
  \centering
  \includegraphics[width=\linewidth]{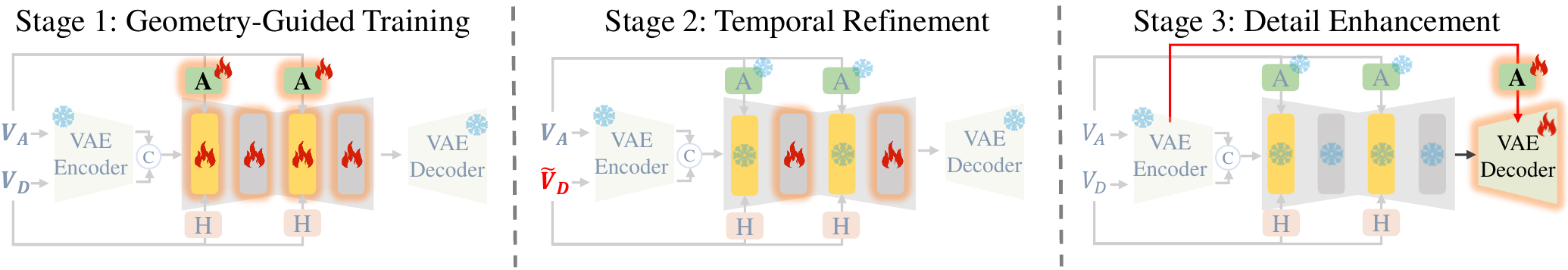}
    \vspace{-5mm}
  \caption{
   \textbf{Progressive Training Strategy}: Stage 1: Train the whole U-Net and adapters. Stage 2: Refine the temporal block with the disturbance. Stage 3: Fine-tuning VAE decoder. We desaturated the colors in the same areas.
  }
  \label{fig: progressive}
  \vspace{-4mm}
\end{figure}

\subsection{MPI Spatial Blocks}
To incorporate geometry into feature processing, we introduce MPI Attention, a gated attention mechanism inspired by~\citep{li2023gligen}, integrated into the U-Net’s spatial blocks, referred to in \cref{fig: pipeline}(b) as MPI spatial blocks. Let $\mathbf{Q}=\{\mathbf{Q}_1,\dots,\mathbf{Q}_i\}$ be query tokens from the current block, $\mathbf{V}_A$ be visual tokens from the input video, and $\gamma$ a learnable gate (initialized to zero). We modulate queries with $\mathbf{K}$ and guide attention with the focal-plane–adapted MPI mask:
\begin{equation}
\label{equ:gate-attention}
\hat{\mathbf{Q}}=\mathbf{Q}+\tanh(\gamma)\cdot
\mathrm{TS}\left(
\mathrm{Attn}\big([\mathbf{Q}+\Phi_M(E(\mathbf{K})),\Phi_A(\mathbf{V}_A)],\bar{\mathcal{M}}\big)
\right),
\end{equation}
where $E(\mathbf{K})$ is a Fourier embedding~\citep{mildenhall2021nerf} of $\mathbf{K}$, $\Phi_M(\cdot)$ and $\Phi_A(\cdot)$ are lightweight MLPs, $\mathrm{TS}$ is a token-selection operator, and $\bar{\mathcal{M}}=[\mathbf{1},\mathcal{M}]$ augments the mask with a global token. We inject \emph{near-focus} masks in shallow layers to refine local transitions and \emph{wider-interval} masks in deeper layers for global context; mask resolutions are aligned to each block by bilinear interpolation. This design steers attention toward focus-relevant regions while maintaining global structure, enabling geometry consistent, contour-aligned bokeh.

\subsection{Progressive Training Strategy}
\label{sec: Progressive Training Strategy}

Although the one-step video diffusion model conditioned on MPI priors provides strong controllability, training such a model directly is challenging. In practice, three issues must be addressed: (i) maintaining temporal consistency across frames, (ii) ensuring robustness against imperfect depth estimation, and (iii) preserving high-frequency details that are often lost during encoding. To systematically tackle these challenges, we adopt a three-stage progressive training strategy, as illustrated in \cref{fig: progressive}.

\textbf{Stage 1: Geometry-Guided Training.}  
We begin by fine-tuning the MPI spatial block, temporal block, and adapters on clean data. Training without stochastic noise encourages the network to fully exploit the focal-plane-adapted MPI mask, enabling it to learn spatially accurate, depth-aware blur effects. This stage allows the model to establish a reliable foundation for spatial geometry and to build initial temporal consistency across video frames.

\textbf{Stage 2: Temporal Refinement with Disparity Noise.}  
A key difficulty in video bokeh is robustness to imperfect or noisy depth estimation, especially around focal boundaries. To address this, we freeze the spatial MPI block and train only the temporal modules on longer sequences while injecting targeted disparity perturbations. Specifically, we apply elastic transforms~\citep{info11020125} to distort depth locally, add Perlin noise~\citep{perlin1985image} to simulate natural inconsistencies, and use morphological operations to exaggerate boundary transitions. This perturbation setting prevents the spatial modules from being overwhelmed while teaching the temporal block to tolerate depth noise and reduce flickering. As a result, the model develops longer temporal memory and stronger resilience to real-world variations.

\textbf{Stage 3: Detail Enhancement via VAE Decoder.}  
Even with stable geometric and temporal modeling, fine details may be lost due to VAE-based compression. In the final stage, we fine-tune the VAE decoder and its adapter using clean data, introducing a skip connection from the encoder to reuse spatially rich features~\citep{parmar2024one}. To emphasize texture fidelity, we combine an image-space $L_1$ loss with a gradient-based texture loss $\mathcal{L}_t$:
\begin{equation}
\label{equ:texture}
\mathcal{L}_t 
= \sum_{x,y} \left[ 
\left( \nabla_x \boldsymbol{\hat{V}}_B(x,y) - \nabla_x \boldsymbol{V}_B(x,y) \right)^2 
+ \left( \nabla_y \boldsymbol{\hat{V}}_B(x,y) - \nabla_y \boldsymbol{V}_B(x,y) \right)^2 
\right],
\end{equation}
where $\mathbf{\hat{V}}_B$ and $\mathbf{V}_B$ denote the predicted and ground truth frames. This loss encourages sharper edges and more realistic textures, thereby enhancing the perceptual quality of the bokeh effect.

Through this progressive strategy, the model gradually learns spatial geometry, temporal robustness, and detail recovery, yielding temporally coherent, depth-aware, and visually realistic video bokeh.

\subsection{Weighted Overlap Inference Strategy}
\label{sec:Weighted Overlap Inference Strategy}

Naively splitting a video into overlapping segments and processing each segment independently can result in noticeable temporal discontinuities at the segment boundaries. These discontinuities often manifest as flickering, misaligned blur transitions, or mismatched bokeh effects, disrupting the smoothness and continuity of the video. 
To address this, we propose a weighted overlapping inference strategy (WOIS).
Specifically, we divide the input video into $P$ overlapping segments, each containing $2L$ frames, with adjacent segments overlapping by exactly $L$ frames. For the $\text{j}$-th frame in the overlap region, denoted as $\hat{V}_B^i[j]$, we compute the final output by combining adjacent frames from overlapping segments using a weighted average:
\begin{equation}
\label{equ:overlap}
\boldsymbol{\tilde{V}}_B^i[j] 
= \gamma_j \, \boldsymbol{\hat{V}}_B^i[j] 
+ \left(1 - \gamma_j\right) \boldsymbol{\hat{V}}_B^{i+1}[j+L],
\quad i \in \{1, 2, \ldots, P\},
\end{equation}
where the weighting factor $\gamma_j$ is defined as:
\begin{equation}
\label{equ:blend factor}
\gamma_j = \frac{1}{2} \left( 1 + \cos( \frac{\pi j}{L} ) \right),
\end{equation}
which smoothly blends the frames at the boundaries, ensuring a gradual transition between segments. The cosine-based weighting function assigns higher weights to the central frames of each segment, reducing the influence of the boundary frames and minimizing visual artifacts at the segment edges.
This strategy enables our framework to process videos of arbitrary length, providing a robust solution to the problem of temporal flickering and boundary artifacts in video bokeh generation.

\section{Experiments}
\subsection{Implementation Details}
\label{sec: implement details}
\textbf{Dataset}. Currently, paired datasets for all-in-focus and bokeh videos are lacking. Existing computational bokeh datasets~\citep{ignatov2020rendering,seizinger2025bokehlicious,dutta2021stacked} primarily consist of image pairs but lack temporal consistency, and previous video-based works~\citep{luo2024video} often omit foreground motion. To address this, we build upon prior work~\citep{bokehme++} and adopt a synthetic approach to generate paired all-in-focus and bokeh video sequences.
For accurate foreground extraction, we use objects from the video matting dataset~\citep{lin2021real}, isolating them using the alpha channel for precise segmentation. To augment the dataset, we also incorporate the image matting dataset~\citep{li2022bridging} and collect 1,300 background images from the intern dataset and background dataset~\citep{lin2021real}. In each video, we randomly select background and foreground clips, simulating real-world camera adjustments such as focal plane and aperture changes. The foreground objects are moved along random 3D trajectories, each containing 25 frames. All videos in the dataset have a resolution of $1024 \times 576$ pixels. Using a ray-tracing-based method~\citep{peng2022mpib}, we generate accurate bokeh effects, ensuring temporal coherence across frames.

\textbf{Training and Inference}. In this work, we use SVD~\citep{blattmann2023stable} as our base model. During training, the process is divided into three stages. Stage 1: We train using 4-frame video sequences with the Adam optimizer at a learning rate of 1e-5. Stage 2: We train with 8-frame video sequences, using a learning rate of 5e-6 and a depth perturbation probability of 0.5. Stage 3: We fine-tune the VAE decoder using the same learning rate as in Stage 2. For all stages, the video resolution is set to 
$1024 \times 576$ with a batch size of 1, and training is performed across 4 Nvidia H800 GPUs.
During inference, we use the weighted overlap inference strategy where long videos are divided into clips of 8 frames, with each clip having a 4-frame overlap.

\textbf{Test Setting}. We validate our model on both synthetic and real-world datasets. First, we synthesize a test set of 200 videos with varying bokeh strength and focus planes. We report the following metrics: PSNR and SSIM~\citep{wang2004ssim} for image fidelity; VFID~\citep{wang2018video} with I3D features~\citep{carreira2017i3d} (denoted VFID-I) and FVD~\citep{ge2024content} for video quality; and the relation metric~\citep{luo2024video} (denoted RM) for temporal consistency, which computes pixel-wise differences between adjacent frames. Additionally, we calculate flow difference (FD) between predicted and ground truth frames using RAFT~\citep{teed2020raft}. For real-world scenarios, we design VEPI, inspired by prior work~\citep{joseph2017edge}, to assess the model’s ability to preserve detail at the edges of the focused subject. We apply the VEPI metric to the DAVIS dataset~\citep{Perazzi_CVPR_2016}, with implementation details described in \cref{sec:VEPI Metric}. We evaluate 25-frame videos and compute the mean inference time.

\begin{table}[!t]
    \centering
    \scriptsize
    \vspace{-2mm}
    \caption{Quantitative comparison of Any-to-Bokeh. The best metric scores in each column are marked in \textbf{bold} for clarity.``$\downarrow$'' or ``$\uparrow$'' indicate lower or higher values are better.}
    \label{tab:baselines}
    \vspace{-2mm}
    \resizebox{\linewidth}{!}{
    \begin{tabular}{lcccccccc}
    \toprule[1pt]
        \textbf{Method}
        & \textbf{FD}$\downarrow$
        & \textbf{RM}$\downarrow$
        & \textbf{VFID-I}$\downarrow$
        & \textbf{FVD}$\downarrow$
        & \textbf{SSIM}$\uparrow$
        & \textbf{PSNR}$\uparrow$
        & \textbf{VEPI}$\uparrow$
        & \textbf{Time}$\downarrow$ \\
\midrule 
 DeepLens~\citep{deeplens2018} & 1.162 & 0.030 & 16.042 & 125.338 & 0.819 & 24.574 & 0.715 & 0.226 \\
 BokehDiff~\citep{zhu2025bokehdiff} & 0.660 & 0.021 & 7.395 & 65.678 & 0.834 & 27.525 & 0.859 & 0.799 \\
 BokehMe~\citep{peng2022bokehme} & 0.536 & 0.013 & 8.633& 39.102 & 0.936 & 27.992 & 0.937 & \textbf{0.103} \\
 Dr.Bokeh~\citep{sheng2024dr} & 0.522 & 0.011 & 6.097& 32.710 & 0.950 & 31.273 & 0.863 & 2.729 \\
 MPIB~\citep{peng2022mpib} & 0.481 & 0.011 & 5.444 & 35.766 & 0.950 & 31.390 & 0.921 & 0.521 \\
 \textbf{Any-to-Bokeh} & \textbf{0.431} & \textbf{0.007} & \textbf{1.479} & \textbf{9.005} & \textbf{0.974} & \textbf{38.899} & \textbf{0.944} & 0.363 \\
\bottomrule[1pt]
\end{tabular}}
\end{table}

\subsection{Results on Test Dataset}
To verify the performance of our proposed method, we compare it with four existing computational bokeh methods: DeepLens~\citep{deeplens2018}, BokehMe~\citep{peng2022bokehme}, Dr.Bokeh~\citep{sheng2024dr},  MPIB~\citep{peng2022mpib}, and BokehDiff~\citep{zhu2025bokehdiff}. Specifically, to compare with traditional MPI-based bokeh methods, MPIB and Dr.Bokeh use the conventional approach, which discretize the scene into front-to-back layers with a fixed depth value. Only one prior work~\citep{luo2024video} addresses video bokeh, but it is not publicly available for comparison.

\begin{figure}[!t]
  \centering
  \includegraphics[width=\linewidth]{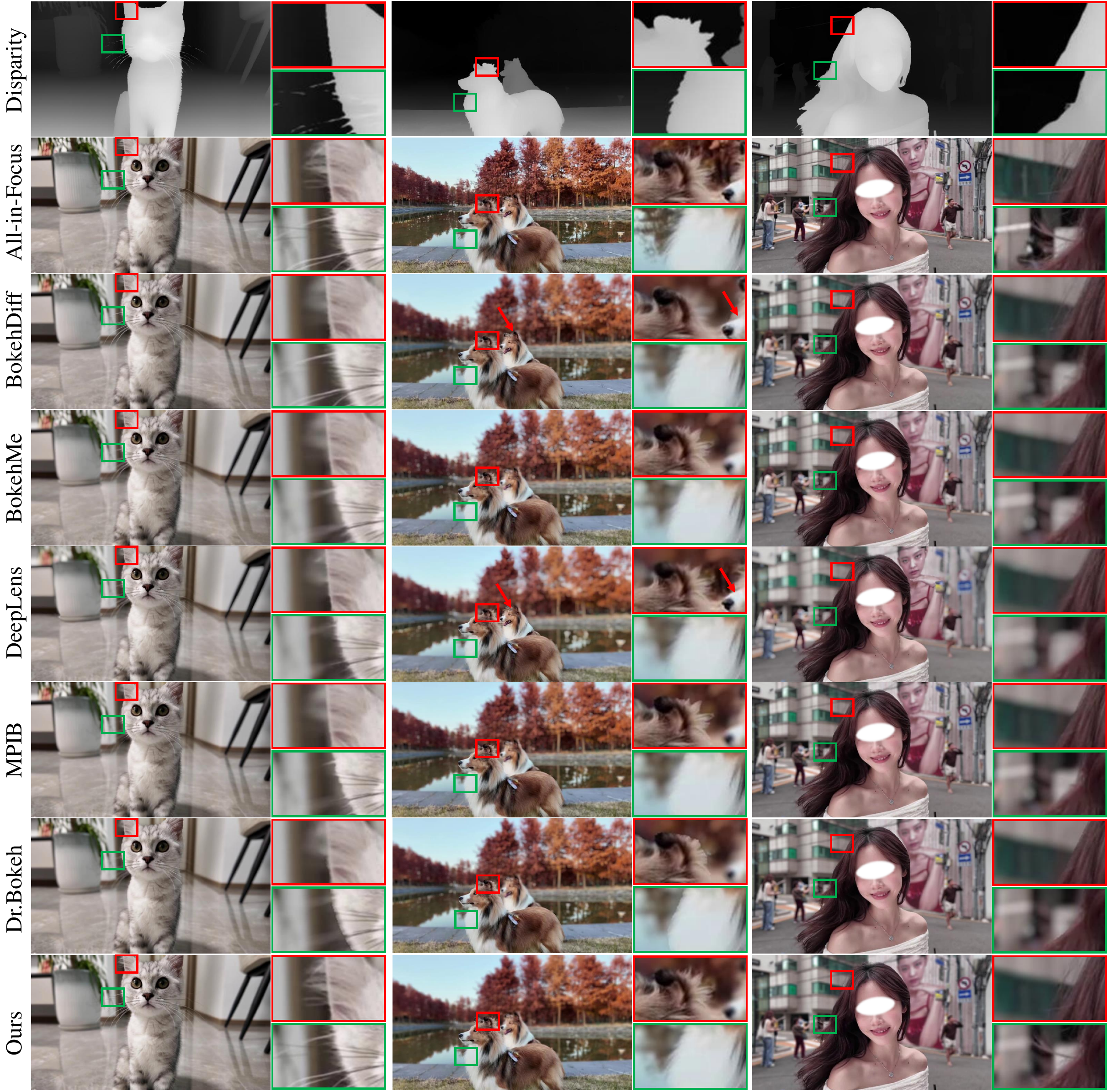}
    \vspace{-5mm}
  \caption{Qualitative Results on Real-World Video frames. To highlight the differences, we zoom in on the red and green regions. Red arrows indicate incorrectly focused areas.
  }
  \label{fig: compare_testset}
  \vspace{-5mm}
\end{figure}

\begin{figure}[!t]
  \centering
  \includegraphics[width=0.85\linewidth]{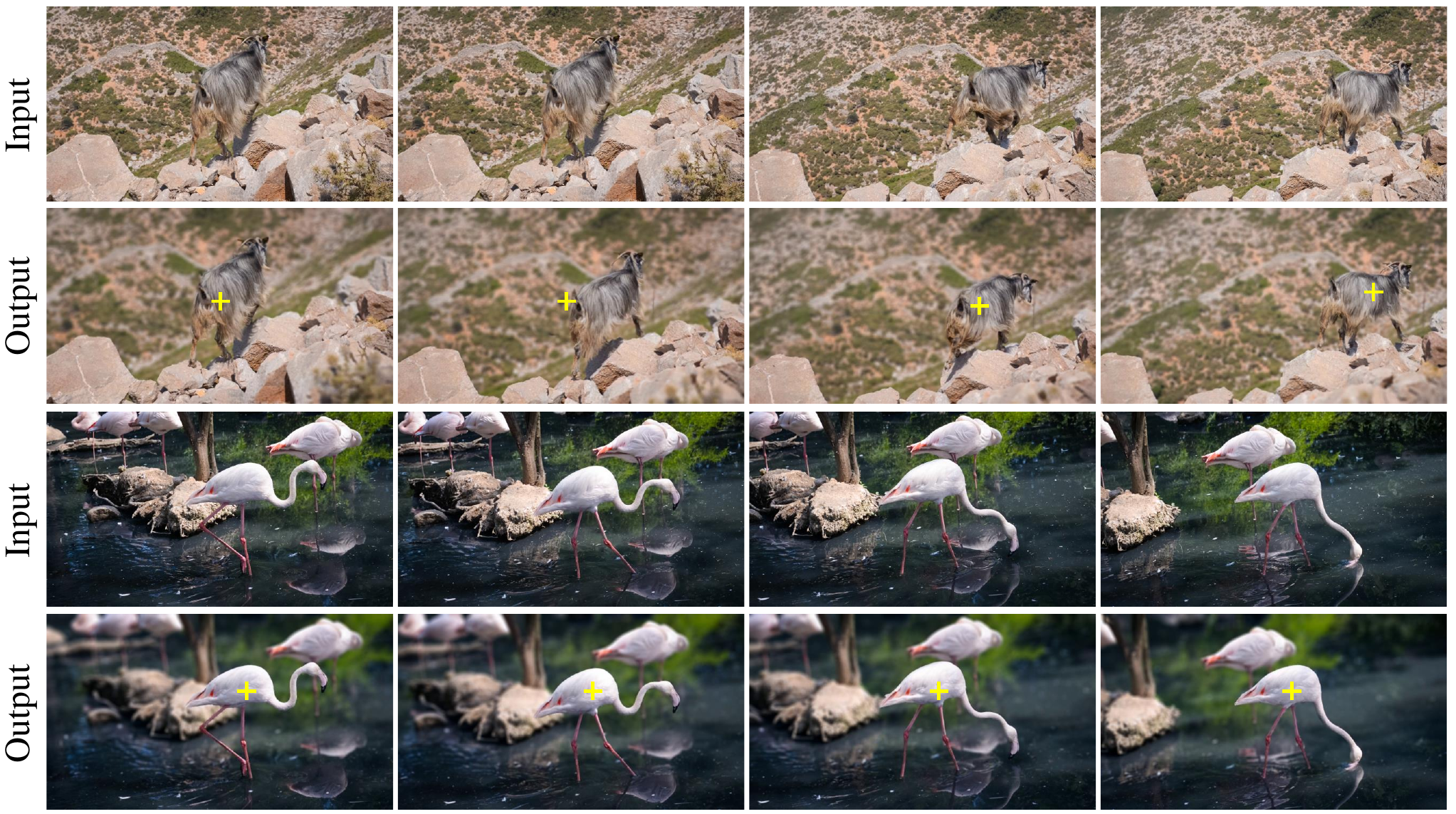}
    \vspace{-3mm}
  \caption{Visualization of generated bokeh effects on DAVIS dataset. The yellow cross represents the focus subject. Please zoom in to see the details.
  }
  \label{fig: visual_res}
  \vspace{-6mm}
\end{figure}

\textbf{Quantitative Results}. As shown in \cref{tab:baselines}, our method consistently outperforms all other approaches across all metrics. Specifically, Any-to-Bokeh achieves the lowest FD and the best RM, demonstrating superior temporal consistency. 
Furthermore, Any-to-Bokeh surpasses the baselines in video quality, attaining the lowest VFID-I and FVD, along with the highest SSIM and PSNR, demonstrating its capability to produce high-quality image fidelity and video bokeh. 
In real-world scenarios, our approach achieves the highest VEPI, demonstrating better blur transitions. Furthermore, compared to traditional MPI methods, our approach exhibits an efficiency advantage in inference time.

\textbf{Qualitative Results}. To further highlight the visual advantages of our approach, we present three examples in \cref{fig: compare_testset}. Any-to-Bokeh is able to preserve detailed foreground textures, such as animal fur and flowing hair, even under poor depth prediction conditions. 
Additionally, it produces bokeh effects that adhere to optical principles across different depth planes, such as the cat’s tail in the first column and the dog in the second column at a farther distance. 
Traditional MPI-based methods, such as MPIB~\citep{peng2022mpib} and Dr.Bokeh~\citep{sheng2024dr}, struggle to blend different depth planes naturally, resulting in color bleeding, as seen in the dog’s ear in the second column. BokehDiff~\citep{zhu2025bokehdiff} and Deeplens~\citep{deeplens2018} can preserve some foreground details, but produce bokeh results that conflict with depth, such as the dog in the second column marked with a red arrow. Moreover, BokehDiff consistently exhibits loss of fine details in its results. BokehMe~\citep{peng2022bokehme} heavily relies on accurate depth information and fails to maintain fur details, as seen in the in-focus subject in the first and third columns when depth is inaccurate. Additional examples are provided in \cref{fig: more_compare}, \cref{fig: more_compare2}, \cref{fig: video_res}, and \cref{fig: video_res2}.

\begin{wraptable}{r}{0.4\textwidth}
    \centering
    \scriptsize
    \vspace{-4mm}
    \caption{Results on human preference.}
    \label{tab:userstudy}
    \vspace{-3mm}
    \resizebox{1.0\linewidth}{!}{
    \begin{tabular}{ccccccc}
    \toprule
         \textbf{Baseline}  
        & \textbf{Preference} \\
    \midrule 
    Ours vs. DeepLens & 96.9\% / 3.1\%  \\
    Ours vs. BokehMe & 77.1\% / 22.9\%  \\
    Ours vs. MPIB & 62.9\% / 37.1\%  \\
    Ours vs. Dr.Bokeh & 77.8\% / 22.2\%  \\
    Ours vs. BokehDiff & 75.7\% / 24.3\%  \\
    \bottomrule
    \end{tabular}}
    \vspace{-3mm}
\end{wraptable}

\textbf{User Study on Real-world Videos}. We test on the DAVIS~\citep{Perazzi_CVPR_2016} dataset. As shown in \cref{fig: visual_res}, our method generates bokeh effects that follow optical principles and maintain strong temporal consistency. We conducted a user study on them to better evaluate different methods from a subjective perspective. We randomly selected 20 videos from the DAVIS dataset and rendered them using various methods. Each participant viewed two videos at a time: one generated by our method and one from a randomly selected baseline. The videos were presented in random order. Participants were asked to choose the method that produced the most aesthetically pleasing bokeh effect based on their personal preference. If they found it difficult to decide, they were allowed to skip making a selection. The user study involved 51 participants and provided 1020 ratings across these video sets. The results in \cref{tab:userstudy} indicate that our method was preferred over the others, demonstrating a higher human preference for the bokeh effects generated by our approach.

\begin{table}[t]
    \centering
    \scriptsize
    \caption{Ablation study of Any-to-Bokeh module. ``MPI'': MPI spatial block. ``OS'': one-step inference schedule. ``WOIS'': weighted overlap inference strategy. ``TR'':  temporal refinement.}
    \vspace{-2mm}
    
    \label{tab:ablation}
    \resizebox{1.0\linewidth}{!}{
    \begin{tabular}{c|ccccccccccc}
    \toprule
        Variants
        & \textbf{MPI}
        & \textbf{OS} 
        & \textbf{WOIS} 
        & \textbf{TR} 
        & \textbf{FD}$\downarrow$
        & \textbf{RM}$\downarrow$
        & \textbf{VFID-I}$\downarrow$
        & \textbf{FVD}$\downarrow$
        & \textbf{SSIM}$\uparrow$
        & \textbf{PSNR}$\uparrow$ \\
\midrule 
 \#1 & \checkmark & \checkmark & \checkmark & \checkmark & 0.517 & 0.013 & 3.865 & 18.922 & 0.907 & 32.250 \\
 \#2 & \checkmark & \checkmark & \checkmark & - &  0.540 & 0.013 & 4.209 & 20.743 & 0.905 & 32.035 \\
 \#3 & \checkmark & \checkmark & - & - & 0.551 & 0.013 & 4.521 & 21.941 & 0.905 & 31.936 \\
 \#4 & - & \checkmark & - & - & 0.573 & 0.013 & 4.930 & 23.828 & 0.903 & 31.551 \\
 \#5 & \checkmark & - & - & - & 0.791 & 0.014 & 8.912 & 68.910 & 0.878 & 29.309 \\
\bottomrule
\end{tabular}}
\vspace{-5mm}
\end{table}

\vspace{-1mm}
\subsection{Ablation Studies}
\vspace{-1mm}
We analyze the impact of each component in the Any-to-Bokeh framework on the synthetic test dataset (results in \cref{tab:ablation}), evaluate the VAE’s contribution to video detail quality (as shown in \cref{tab:ablation_vae}), and assess the effect of temporal refinement on model robustness (as shown in \cref{tab:ablation_robust}).

\textbf{MPI Spatial Block Ablation}. Lines 3 and 4 in \cref{tab:ablation} show the ablation results for the MPI block. First, we remove the MPI module and use the original attention block, adding the blur strength ($K$) embedding directly to the original SVD embedding. From the VFID and FVD metrics, we observe that our model benefits significantly from the MPI blocks, leading to better video quality. Additionally, the MPI block enhances temporal consistency and single-frame quality, which are crucial for the bokeh rendering task. Additionally, we report results for two simple alternatives to MPI Attention, further demonstrating its effectiveness. Please refer to \cref{sec:More Ablation MPI Attention}.

\textbf{Effectiveness of One-Step Inference}. We add noise to the input video features and retrain a multi-step denoising model, comparing its performance with the one-step inference schedule. As shown in the 3rd and 5th rows of \cref{tab:ablation}, the one-step inference schedule substantially improves all metrics, highlighting its ability to achieve more temporally consistent and higher-quality bokeh effects.

\textbf{Effectiveness of Weighted Overlap Inference Strategy (WOIS)}. By incorporating WOIS, our model achieves improved temporal consistency, with the flow difference (FD) decreasing from 0.551 to 0.540. Additionally, higher FVD and VFID scores indicate enhanced video quality. Furthermore, the PSNR also shows an increase, reflecting improved single-frame fidelity. More detail are provided in \cref{sec:ablation Weighted Overlap Inference Strategy}.

\begin{table}[ht]
\vspace{-1mm}
    \noindent 
    \begin{minipage}{0.38\linewidth}
        \caption{Ablation study on FV.}
        \vspace{-3mm}
        \label{tab:ablation_vae}
        \scriptsize
        \resizebox{1.0\linewidth}{!}{
        \begin{tabular}{ccccccc}
            \toprule
            \textbf{FV} & \textbf{FVD}$\downarrow$ & \textbf{SSIM}$\uparrow$ & \textbf{PSNR}$\uparrow$ \\
            \midrule 
            \checkmark & 9.005 & 0.974 & 38.899 \\
             - & 18.922 & 0.907 & 32.250 \\
            \bottomrule
        \end{tabular}}
    \end{minipage}
    \hspace{0.2cm} 
        \begin{minipage}{0.6\linewidth}
        \centering
        \scriptsize
        \caption{Ablation study on TR under noisy depth.}
        \vspace{-3mm}
        \label{tab:ablation_robust}
        \resizebox{1.0\linewidth}{!}{
        \begin{tabular}{ccccccc}
            \toprule
            \textbf{TR} & \textbf{FD}$\downarrow$ & 
            \textbf{VFID-I}$\downarrow$ &
            \textbf{FVD}$\downarrow$ &
            \textbf{SSIM}$\uparrow$ & \textbf{PSNR}$\uparrow$ \\
            \midrule 
            \checkmark & 0.531 & 3.880 & 19.648 & 0.906 & 32.182 \\
            - & 0.566 & 4.442 & 22.452 & 0.904 & 31.890 \\
            \bottomrule
        \end{tabular}}
    \end{minipage}
    \vspace{-3mm}
\end{table}

\textbf{Effectiveness of Progressive Training Strategy}. We begin the ablation study by evaluating the full model. In the second row in \cref{tab:ablation}, we remove the temporal block refinement (TR) from stage 2, resulting in a decrease in temporal consistency (FD: 0.517 vs. 0.540). This highlights the importance of temporal block refinement in maintaining temporal coherence across video frames. As shown in \cref{tab:ablation_vae}, we alleviated the loss of high-frequency information by fine-tuning the VAE (FV), which effectively improves both single-frame consistency and overall video quality.

\textbf{Ablation Study on Robustness.} To test the contribution of TR to robustness, we introduce perturbations to the disparity in the test dataset using elastic transform~\citep{info11020125}, Gaussian blur, and morphological transformations. As shown in the last two rows in \cref{tab:ablation_robust}, TR leads to improvements across all metrics, with particularly noticeable gains in temporal consistency (FD) and video quality (VFID-I and FVD). These results demonstrate that training the temporal blocks with noisy data during the TR stage effectively enhances the model's robustness. Furthermore, \cref{sec:Quantitative Comparison Under Noisy Depth} presents a comparison of our method with baselines under noisy depth conditions, emphasizing the robustness of our approach, especially in preserving temporal consistency.

\vspace{-1mm}
\section{Conclusions}
In this work, we propose the first one-step diffusion framework for controllable video bokeh. By incorporating an MPI-guided conditioning mechanism that injects explicit scene geometry, our method achieves higher-quality and more temporally consistent bokeh effects. Additionally, we introduce a progressive training strategy that enhances robustness and detail preservation, significantly improving bokeh quality.
We hope our findings inspire further exploration of optical phenomena in editing models, driving advancements in their application to content creation and visual effects. By better understanding these phenomena, we aim to enhance the realism and flexibility of future editing models, enabling more creative possibilities for the industry.




\bibliography{any2bokeh}
\bibliographystyle{iclr2026_conference}
\newpage

\appendix
\section{Appendix}

\begingroup
\setcounter{tocdepth}{3} 
{\hypersetup{linkcolor=black}\localtableofcontents} 
\endgroup

\clearpage
In the appendix, we begin with the LLM usage statement, followed by a detailed description of our dataset, user study, and the VEPI metric. We then report additional ablation studies, covering MPI Attention and the weighted overlap inference strategy. We also present more comparison results, including qualitative comparisons under noisy inputs and additional visualizations, to further demonstrate the superiority of our approach. Finally, we provide computational comparisons with baseline methods.

\subsection{Usage of LLM}
\label{sec:Usage of LLM}
In this study, a large language model (LLM) was used to polish the writing and elevate the overall textual quality.

\subsection{The Details of Dataset}
\label{sec:The Details of Dataset}

\begin{figure}[htbp]
  \centering
  \includegraphics[width=\linewidth]{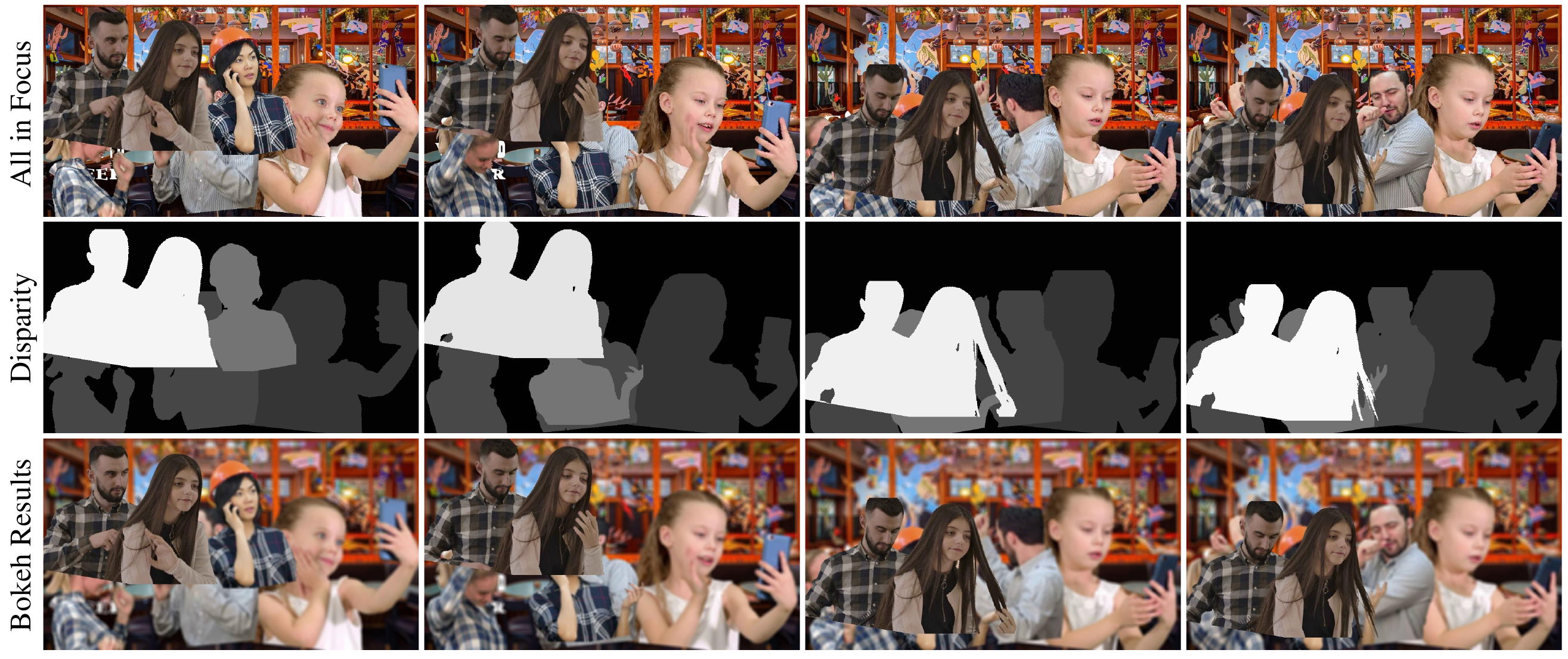}
    \vspace{-5mm}
  \caption{Example of synthetic datasets, we randomly define the focal plane, the position of each foreground, and the blur intensity.
  }
  \label{fig: dataset}
  \vspace{-2mm}
\end{figure}

As mentioned on \cref{sec: implement details}, we use objects from the video matting dataset~\citep{lin2021real} and image matting dataset~\citep{li2022bridging} and collect 1,300 background images from the intern dataset and background dataset~\citep{lin2021real}. 
Following MPIB~\citep{peng2022mpib}, we use a ray-tracing-based method to generate accurate bokeh effects. Specifically, we assume that the disparities of all images are planar, with their size and position randomly determined. The disparity map $d$ is then set as a plane equation of pixel coordinates $(x, y)$
\begin{equation}
    d = \frac{1 - ax - by}{c},
\end{equation}
where $a,b$, and $c$ are parameters that define the spatial depth relationship between pixels. For each pixel, we sample multiple rays passing through the lens, find the intersection of each ray with the scene, and project this intersection onto the sensor plane to obtain the final render results. 
As shown in \cref{fig: dataset}, for each video, we randomly select background and foreground clips, as well as the focal plane and aperture. The foreground objects are moved along random 3D trajectories, with movement in six dimensions: forward and backward, left and right, up and down. Each video contains 25 frames, and all videos in the dataset have a resolution of $1024 \times 576$ pixels.
To evaluate model performance, we use the same approach to synthesize a test set of 200 videos.

\begin{figure}[!t]
  \centering
  \includegraphics[width=\linewidth]{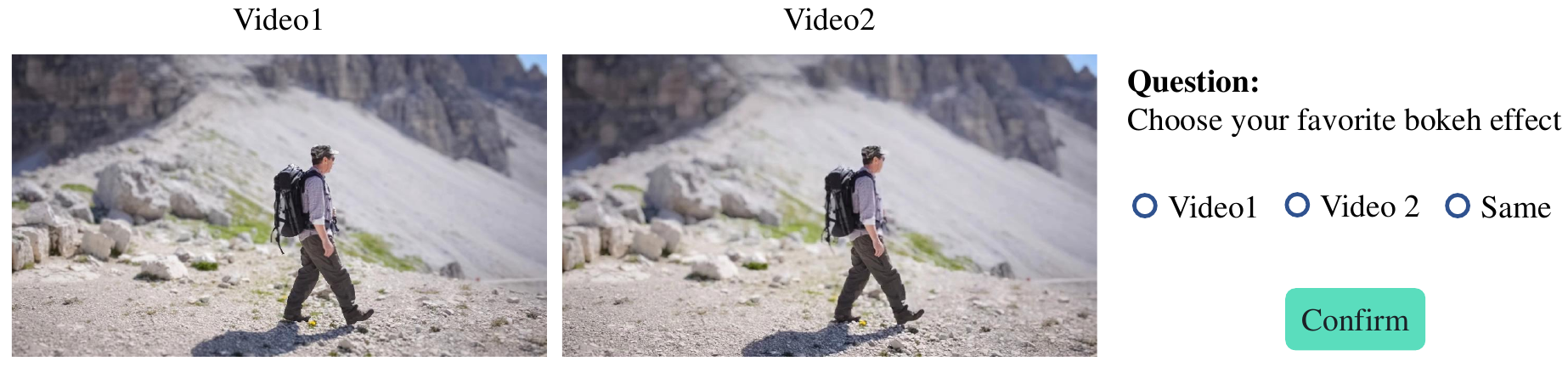}
    \vspace{-5mm}
  \caption{The user study interface, where they were asked to select their preferred videos.
  }
  \label{fig: interface}
  \vspace{-2mm}
\end{figure}

\subsection{User Study Details}
\label{sec:User Study Details}
Given the subjective nature of perceiving video bokeh rendering results, and the absence of ground truth (GT) data for video bokeh, we conducted a user study on real-world videos to evaluate different methods from a subjective perspective. We randomly selected 20 videos from the DAVIS dataset, each with a resolution of 
$1024 \times 576$, featuring subjects such as people, animals, and various other objects. Since ground truth disparity maps were unavailable, we generated disparity maps using a depth prediction model~\citep{chen2025video}.
Different methods were used to render the videos with the same control parameters. During testing, the videos were presented in random order to avoid bias. As shown in the interface (\cref{fig: interface}), participants were asked to select the method that produced the most consistent and aesthetically pleasing bokeh effect. If they found it difficult to decide, they were allowed to skip making a selection. 

\subsection{VEPI Metric Details}
\label{sec:VEPI Metric}
We adapt the Edge Preservation Index (EPI) from prior image-based work~\citep{joseph2017edge}. In VEPI, a Laplacian kernel is applied to generate binary edge maps for both the reference image and the bokeh-rendered image:

\begin{equation}
    \text{VEPI} = \frac{\Gamma(\Delta s - \overline{\Delta s}, \Delta\hat{s} - \overline{\Delta\hat{s}})}{\sqrt{\Gamma(\Delta s - \overline{\Delta s}, \Delta s - \overline{\Delta s})} \cdot \Gamma(\Delta\hat{s} - \overline{\Delta\hat{s}}, \Delta\hat{s} - \overline{\Delta\hat{s}})},
\end{equation}
where
\begin{equation}
    \Gamma(s_1, s_2) = \sum_{i,j \in \text{ROI}} s_1(i,j) \cdot s_2(i,j).
\end{equation}
ROI denotes the unblurred foreground region. $\Delta s(i,j)$ and $\Delta\hat{s}(i,j)$ are the Laplacian-filtered versions of the ROI in the reference $s(i,j)$ and the bokeh rendering $\hat{s}(i,j)$, respectively. Since this metric relies on precise foreground delineation, we employ the DAVIS dataset~\citep{Perazzi_CVPR_2016}, which provides densely annotated video segmentation. We compute VEPI for all videos in the DAVIS dataset, using the provided segmentation masks as the ROI, and report the mean value across all videos as the final result.

\subsection{More Ablation Study}
\label{sec:More Ablation Study}
\subsubsection{MPI Attention}
\label{sec:More Ablation MPI Attention}

\begin{table}[ht]
    \centering
    \caption{Ablation results for MPI attention. ``MPI'' refers to MPI attention, while ``PR'' indicates pretrained SVD weights. }
    \begin{tabular}{ccccccccc}
        \toprule
        Variants &
        \textbf{MPI} & 
        \textbf{PR} & 
        \textbf{FD}$\downarrow$ & \textbf{RM}$\downarrow$ & 
        \textbf{VFID-I}$\downarrow$ & \textbf{FVD}$\downarrow$ & \textbf{SSIM}$\uparrow$ & \textbf{PSNR}$\uparrow$ \\
        \midrule
        \#1 & \checkmark & \checkmark & 0.551 & 0.013 & 4.521 & 21.941 & 0.905 & 31.936 \\
        \#2 & - &  \checkmark & 0.568 & 0.014 & 4.714 & 22.556 & 0.901 & 31.575 \\
        \#3 & \checkmark & -  & 0.586 & 0.014 & 4.537 & 27.743 & 0.893 & 30.988 \\
        \bottomrule
    \end{tabular}
    \label{tab:mpi_ablation}
\end{table}

To validate the role of MPI Attention within the MPI spatial block, we report results for two simple alternatives. First, we discard the guidance of the focal-plane–adapted MPI mask and replace MPI attention with standard self-attention. As shown in the second row of \cref{tab:mpi_ablation}, all metrics decrease, indicating that the focal-plane–adapted MPI mask effectively provides a geometry prior, resulting in temporally consistent and high-quality bokeh. Second, we remove the pretrained SVD weights and train the MPI spatial block from scratch. The results in the third row of \cref{tab:mpi_ablation} demonstrate that the pretrained 3D priors significantly improve model performance.

\subsubsection{Weighted Overlap Inference Strategy}
\label{sec:ablation Weighted Overlap Inference Strategy}

\begin{table}[ht]
    \centering
    \caption{Results for different blending strategy.}
    \begin{tabular}{ccccccc}
        \toprule
        Method & FD$\downarrow$ & RM$\downarrow$ & VFID-I$\downarrow$ & FVD$\downarrow$ & SSIM$\uparrow$ & PSNR$\uparrow$ \\
        \midrule
        Linear & 0.422 & 0.007 & 1.565 & 9.168 & 0.972 & 38.706 \\
        Cosine & 0.431 & 0.007 & 1.479 & 9.005 & 0.974 & 38.899 \\
        \bottomrule
    \end{tabular}
    \label{tab:blend methods}
\end{table}

As introduced in \cref{sec:Weighted Overlap Inference Strategy}, the weighted overlap inference strategy employs a weighting factor for blending overlapping frames. We evaluate two schemes: (i) cosine-based blending, which emphasizes central frames within each segment, and (ii) linear-based blending, which assigns weights linearly as $\gamma_j = \frac{j}{L}$.
As shown in \cref{tab:blend methods}, cosine-based blending provides superior temporal consistency and higher-quality bokeh. Consequently, we adopt the cosine-based blending (\cref{equ:blend factor}) as the weighting factor for overlapping frame fusion.

\subsection{More Comparison Results}
\label{sec:More Comparison Results}

\subsubsection{Quantitative Comparison Under Noisy Depth}
\label{sec:Quantitative Comparison Under Noisy Depth}
To simulate inaccurate depth, we apply elastic transforms, Gaussian blur, and morphological operations to the input depth sequences. In \cref{tab:Noisy Depth}, we compare all baseline methods under these noisy depth conditions. Our approach consistently attains the highest performance across four metrics, exhibiting smaller performance degradation than the baselines. This highlights the robustness of our method, especially in preserving temporal consistency under noisy depth inputs.

\begin{table}[ht]
\centering
\caption{Comparison of baselines under noisy depth. The bold values indicate the best results. Numbers in parentheses indicate the relative performance drop due to noisy depth.}
\label{tab:Noisy Depth}
\begin{tabular}{l|cccccc}
    \toprule
    Method
     & \textbf{FD}$\downarrow$ 
     & \textbf{RM}$\downarrow$ 
     & \textbf{SSIM}$\uparrow$ 
     & \textbf{PSNR}$\uparrow$ \\
    \midrule
    DeepLens & 1.161 (-0.05\%) & 0.030 (-0.00\%) & 0.819 (-0.00\%) & 24.574 (-0.00\%) \\
    BokehDiff & 0.768 (-16.36\%) & 0.021 (-0.00\%) &  0.844 -(1.20\%)  & 27.328 (\textbf{-0.72}\%)  \\
    BokehMe        & 0.564 (-5.15\%)  & 0.014 (-4.78\%) & 0.933 (-0.29\%)  & 27.899 (-0.33\%) \\
    Dr.Bokeh & 0.540 (-3.51\%)  & 0.012 (-5.16\%) & 0.948 (-0.22\%)  & 31.236 (-0.12\%)  \\
    MPIB & 0.540 (-12.22\%) & 0.012 (-10.10\%)  & 0.946 (-0.38\%)  & 30.278 (-3.54\%) \\
    \textbf{Any-to-Bokeh} & \textbf{0.432} (\textbf{-0.23}\%) & \textbf{0.007} (\textbf{-0.00}\%)  & \textbf{0.972} (\textbf{-0.20}\%) & \textbf{38.599} (-0.77\%)\\
    \bottomrule
\end{tabular}
\end{table}

\subsubsection{More Visualization Results}
\label{sec:More Visualization Results}

\begin{figure}[t]
  \centering
  \includegraphics[width=\linewidth]{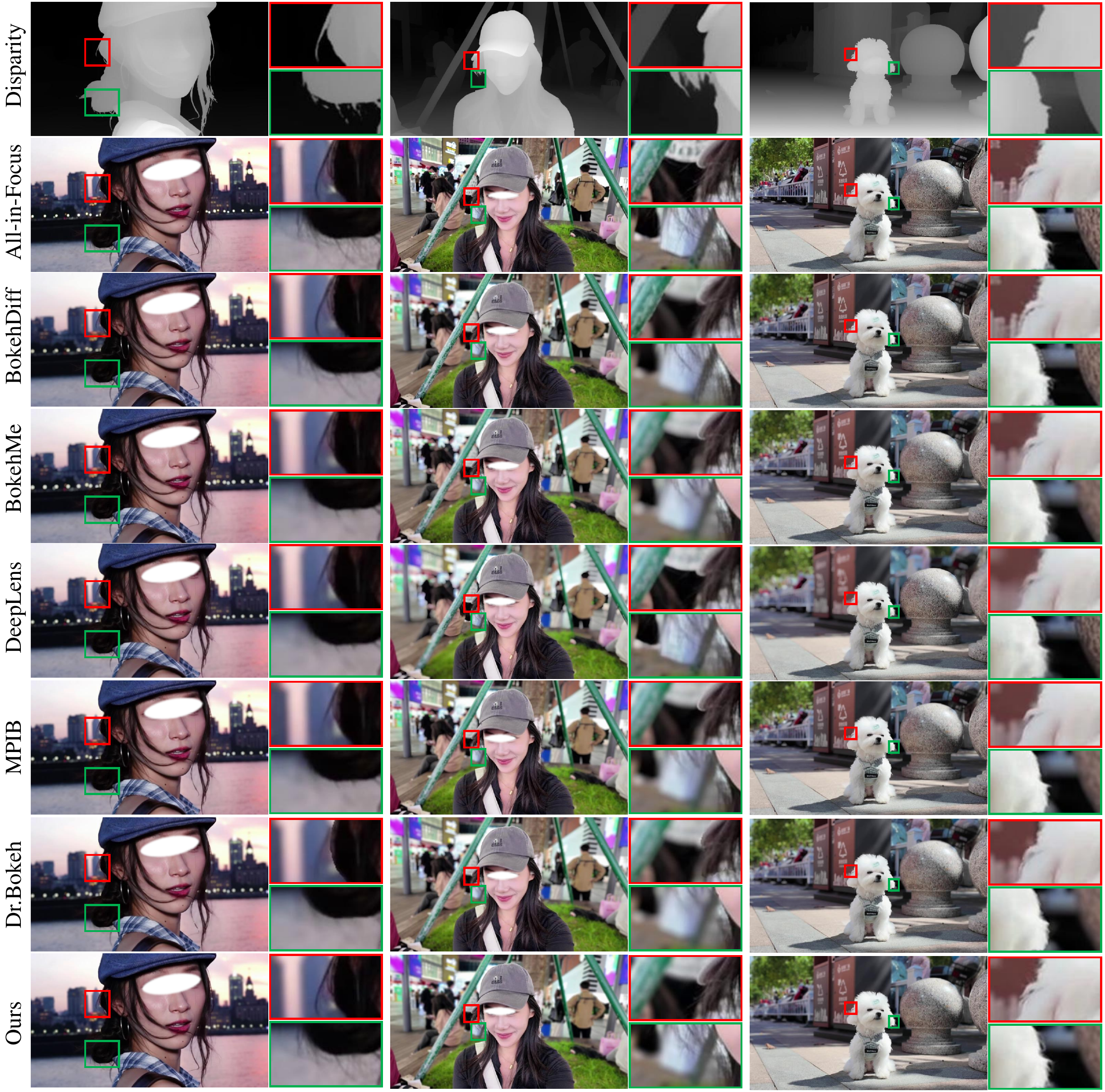}
    \vspace{-5mm}
  \caption{Comparison results with baselines under suboptimal depth inputs. To highlight the differences, we zoom in on the red and green regions.
  }
  \label{fig: more_compare}
  \vspace{-2mm}
\end{figure}

\begin{figure}[t]
  \centering
  \includegraphics[width=\linewidth]{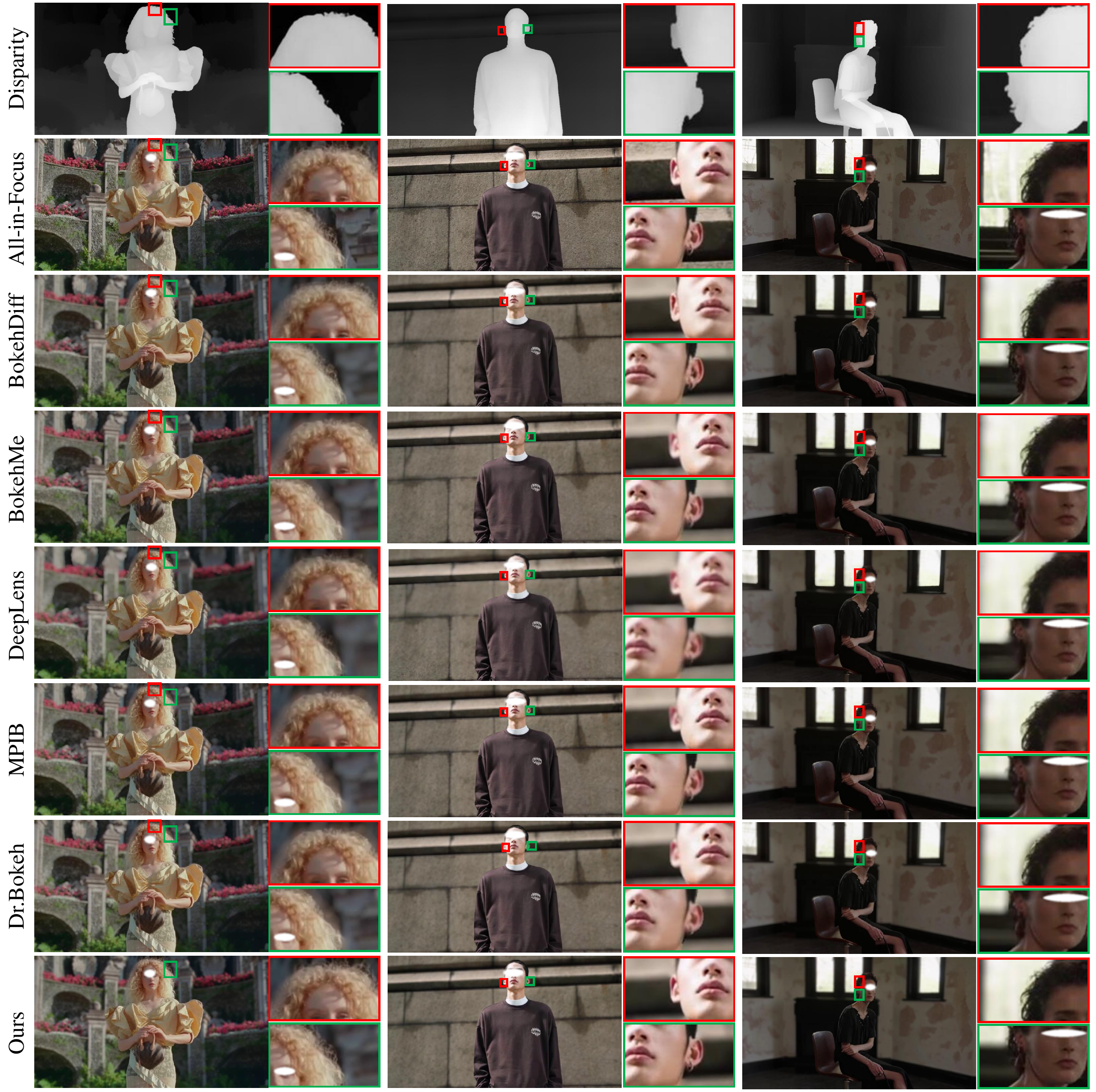}
    \vspace{-5mm}
  \caption{Comparison results with baselines under suboptimal depth inputs. To highlight the differences, we zoom in on the red and green regions.
  }
  \label{fig: more_compare2}
  \vspace{-2mm}
\end{figure}

\textbf{Visualization Results under Suboptimal Depth Inputs}: We present further comparisons between Any-to-Bokeh and baselines to validate its ability to render bokeh under poor depth conditions. In \cref{fig: more_compare} and \cref{fig: more_compare2}, errors in depth prediction are visible along fine edges, such as hair, dog fur, or earrings. BokehMe, DeepLens, MPIB, and Dr.Bokeh rely heavily on accurate depth; in regions with missing or erroneous depth, they fail to capture foreground edges, resulting in degraded bokeh. Although BokehDiff can render some edge textures, it struggles to remain consistent with the input. Our method effectively addresses these challenges, producing spatially accurate, depth-aware blur.

\begin{figure}[t]
  \centering
  \includegraphics[width=\linewidth]{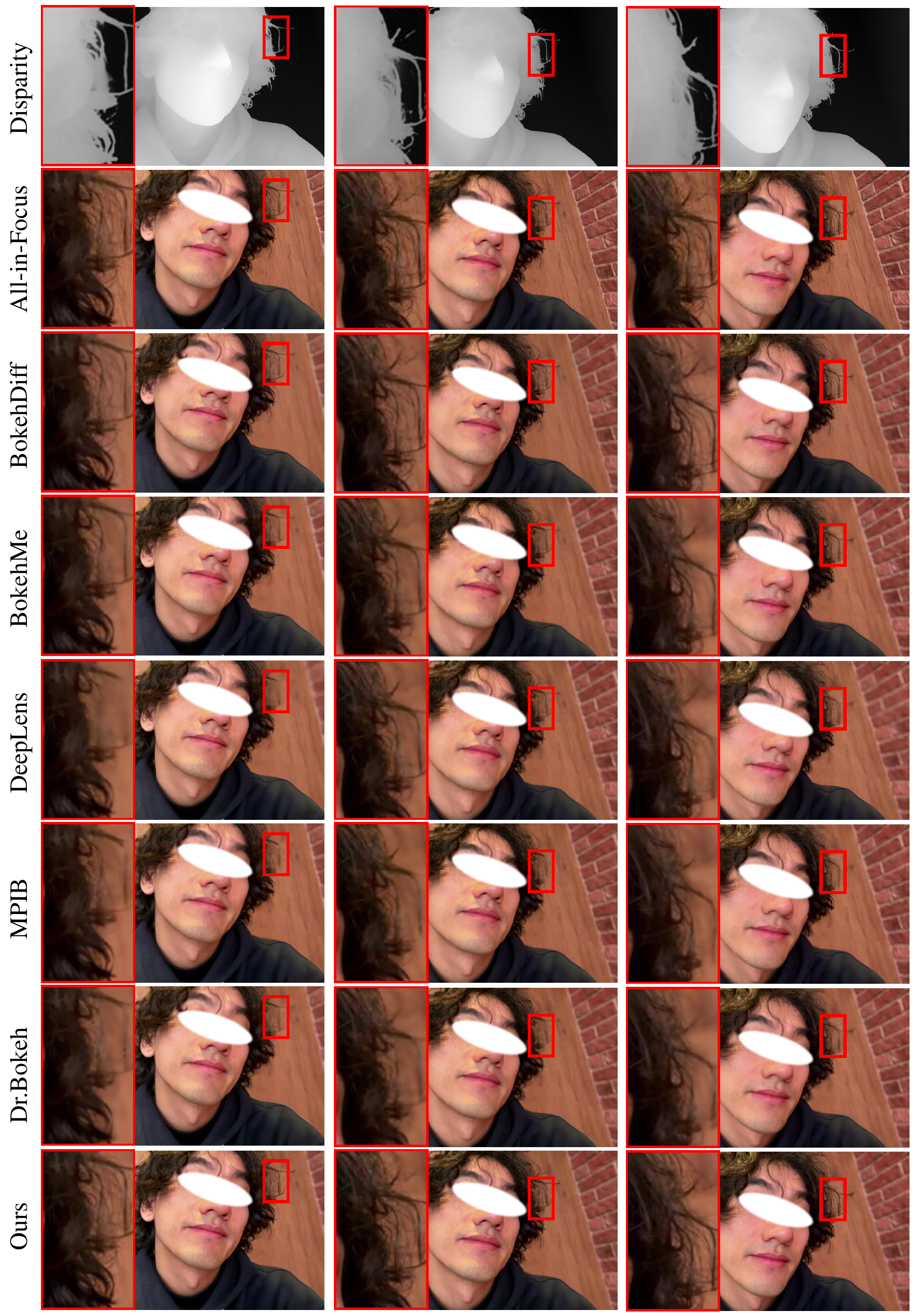}
    \vspace{-5mm}
  \caption{Comparison results with baselines on real-world videos. The area inside the red border is zoomed in to highlight more details. Please zoom in to view them.
  }
  \label{fig: video_res}
  \vspace{-2mm}
\end{figure}

\begin{figure}[t]
  \centering
  \includegraphics[width=\linewidth]{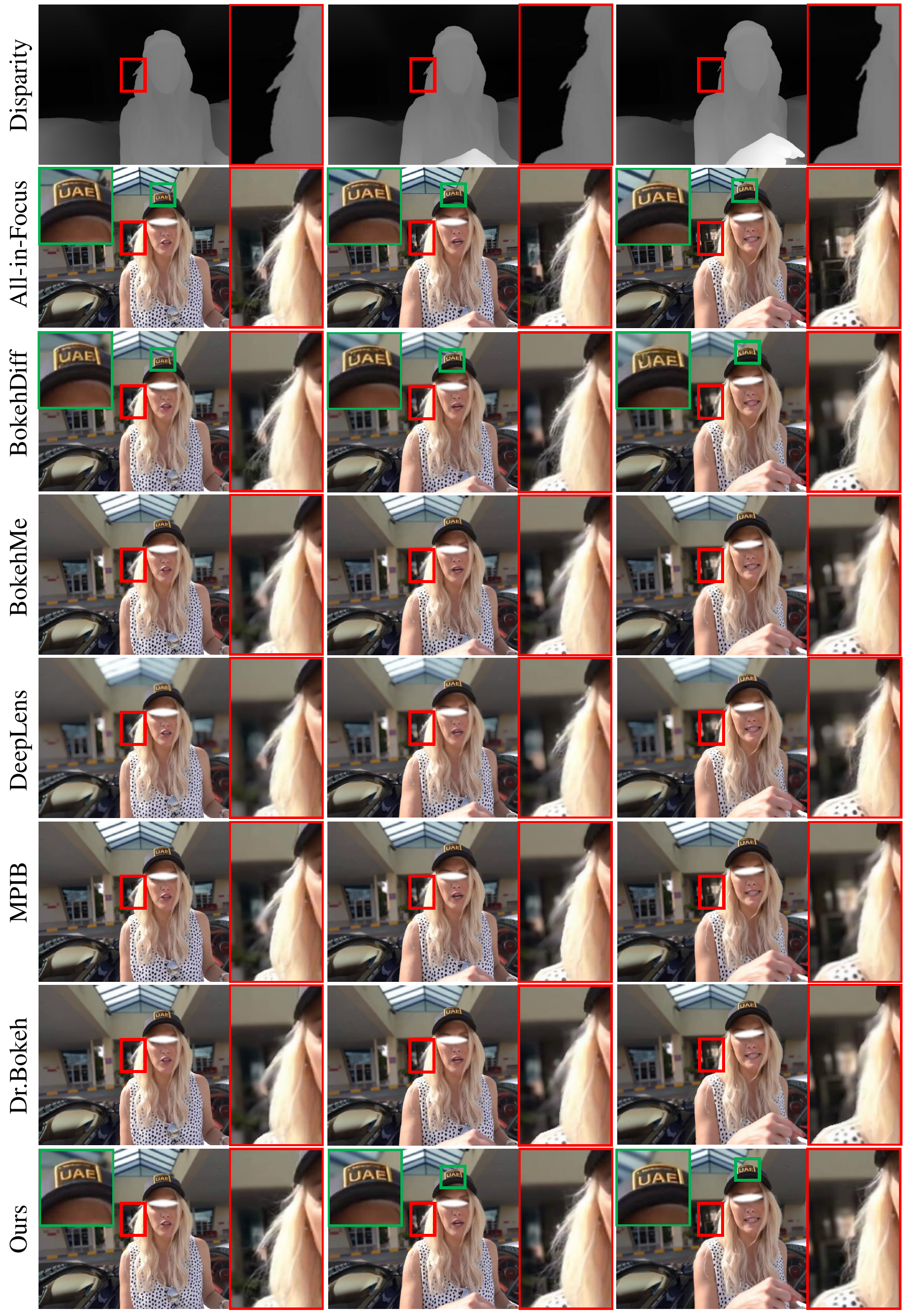}
    \vspace{-5mm}
  \caption{Comparison results with baselines on real-world videos. The area inside the red and green border is zoomed in to highlight more details. Please zoom in to view them.
  }
  \label{fig: video_res2}
  \vspace{-2mm}
\end{figure}

\textbf{Real-World Videos Comparison}: We further compare our model with baselines using two representative examples. As shown in \cref{fig: video_res} and \cref{fig: video_res2}, our method preserves the fine details of the in-focus subject, achieving more visually coherent bokeh. Even under challenging depth conditions, it accurately captures the edges of the subject, enhancing the bokeh realism. In contrast, BokehDiff struggles with the accurate rendering of subject textures, such as the characters on the hat (green box) and hair strands (red box). Furthermore, BokehMe, DeepLens, MPIB, and Dr.Bokeh are unable to preserve hair details within the same focal plane due to poor depth estimation.

\subsubsection{Comparison about Computational}
\label{Comparison about Computational}

\cref{tab:Computational} presents computational comparisons between our method and baselines. BokehMe and Dr.Bokeh use custom CUDA-based renderers, making it difficult to report full FLOPs or parameter counts; for BokehMe, only neural network components are reported (marked with $\ast$). As the baselines are single-frame image bokeh models, we evaluate using 576×1024 resolution single-frame inputs. Measurements are conducted on the same H800 GPU. While our full model employs a larger SVD backbone, end-to-end optimization and a transformer architecture optimized for parallelism result in faster runtime than MPI-based baselines (MPIB and Dr.Bokeh) and BokehDiff.

To explore the parameter-performance trade-off, we introduce \textbf{Ours-tiny}, a lightweight distilled variant trained via knowledge distillation from the full model. It reduces parameters by 6× while maintaining strong performance. Even this distilled model outperforms all baselines, demonstrating an effective balance between efficiency and quality, with future work focused on exploring more lightweight yet high-performing architectures.

\begin{table}[t]
\centering
\caption{Computational comparisons across baselines.}
\resizebox{1.0\linewidth}{!}{
\begin{tabular}{lccccccccc}
\toprule
Method & FD$\downarrow$ & RM$\downarrow$ & VFID-I$\downarrow$ & FVD$\downarrow$ & SSIM$\uparrow$ & PSNR$\uparrow$ & Time (s) & Params (M) & GFLOPs \\
\midrule
BokehMe   & 0.536 & 0.013 & 8.633 & 39.102 & 0.936 & 27.992 & 0.103 & $\ast$1.41  & $\ast$169.71 \\
MPIB      & 0.481 & 0.011 & 5.444 & 35.766 & 0.950 & 31.390 & 0.521 & 27.46       & 1418.50 \\
Dr.Bokeh  & 0.522 & 0.011 & 6.097 & 32.710 & 0.950 & 31.273 & 2.729 & --          & -- \\
BokehDiff      & 0.660 & 0.021 & 7.395 & 65.678 & 0.834 & 27.525 & 0.799 & 2459 & 3430 \\
\midrule
\textbf{Ours}      & 0.431 & 0.007 & 1.479 & 9.005 & 0.974 & 38.899 & 0.094 & 1880 & 3620 \\
\textbf{Ours-tiny} & 0.497 & 0.009 & 2.648 & 16.628 & 0.964 & 36.391 & 0.054 & 319.85 & 586.78 \\
\bottomrule
\end{tabular}}
\label{tab:Computational}
\end{table}

\end{document}